\documentclass{article}

	\PassOptionsToPackage{square,numbers,sort&compress}{natbib}
     \usepackage[final]{neurips_2019}

\usepackage[utf8]{inputenc} % allow utf-8 input
\usepackage[T1]{fontenc}    % use 8-bit T1 fonts
\usepackage{hyperref}       % hyperlinks
\usepackage{url}            % simple URL typesetting
\usepackage{booktabs}       % professional-quality tables
\usepackage{amsfonts}       % blackboard math symbols
\usepackage{nicefrac}       % compact symbols for 1/2, etc.
\usepackage{microtype}      % microtypography
\usepackage{graphicx}
\usepackage{subcaption}
\usepackage{multirow}
\usepackage{textcomp}
\usepackage{natbib}

\usepackage{xargs}                      % Use more than one optional parameter in a new commands
\usepackage[pdftex,dvipsnames]{xcolor}  % Coloured text etc.

\usepackage[colorinlistoftodos,prependcaption,textsize=small]{todonotes}
\newcommandx{\unsure}[2][1=]{\todo[linecolor=red,backgroundcolor=red!25,bordercolor=red,#1]{#2}}
\newcommandx{\change}[2][1=]{\todo[linecolor=blue,backgroundcolor=blue!25,bordercolor=blue,#1]{#2}}
\newcommandx{\info}[2][1=]{\todo[linecolor=OliveGreen,backgroundcolor=OliveGreen!25,bordercolor=OliveGreen,#1]{#2}}
\newcommandx{\improvement}[2][1=]{\todo[linecolor=Plum,backgroundcolor=Plum!25,bordercolor=Plum,#1]{#2}}
\newcommandx{\thiswillnotshow}[2][1=]{\todo[disable,#1]{#2}}

\title{Evaluating robustness of language models for chief complaint extraction from patient-generated text}

\author{ %
   Ilya Valmianski \\
  Medical Informatics Group \\
  Kaiser Permanente \\
  \texttt{Ilya.Valmianski@kp.org} \\ 
  \And 
 Caleb Goodwin \\
 Medical Informatics Group \\
 Kaiser Permanente \\
 \texttt{Joshua.C.Goodwin@kp.org} \\
  \And  
  Ian M. Finn \\
 Medical Informatics Group \\
 Kaiser Permanente \\
 \texttt{Ian.M.Finn@kp.org} \\
  \And 
   Naqi Khan  \\
 Medical Informatics Group 
 \\ Kaiser Permanente \\
 \texttt{Naqi.Khan@kp.org} \\
  \And
   Daniel S. Zisook  \\
 Medical Informatics Group \\
 Kaiser Permanente\\
 \texttt{Daniel.S.Zisook@kp.org} 
}

\begin{document}

\maketitle

\begin{abstract}
	Automated classification of chief complaints from patient-generated text is a critical first step in developing scalable platforms to triage patients without human intervention. In this work, we evaluate several approaches to chief complaint classification using a novel Chief Complaint (CC) Dataset that contains \texttildelow 200,000 patient-generated reasons-for-visit entries mapped to a set of 795 discrete chief complaints. We examine the use of several fine-tuned bidirectional transformer (BERT) models trained on both unrelated texts as well as on the CC dataset. We contrast this performance with a TF-IDF baseline. Our evaluation has three components: (1) a random test hold-out from the original dataset; (2) a "misspelling set," consisting of a hand-selected subset of the test set, where every entry has at least one misspelling; (3) a separate experimenter-generated free-text set. We find that the TF-IDF model performs significantly better than the strongest BERT-based model on the test (best BERT PR-AUC 0.3597±0.0041 vs TF-IDF PR-AUC 0.3878±0.0148, $p=7\cdot 10^{-5}$), and is statistically comparable to the misspelling sets (best BERT PR-AUC 0.2579±0.0079 vs TF-IDF PR-AUC 0.2733±0.0130, $p=0.06$ ). However, when examining model predictions on experimenter-generated queries, some concerns arise about TF-IDF baseline's robustness. Our results suggest that in certain tasks, simple language embedding baselines may be very performant; however, truly understanding their robustness requires further analysis.
\end{abstract}

\section{Introduction}

The first step in many clinical encounters is to obtain the chief complaint, which is a short description of the patient’s reason for seeking medical care. Chief complaint data subserve a myriad of use cases, including syndromic surveillance \citep{brown_ngram_2010, chapman_classifying_2005, chapman_classification_2005, lee_chief_2018}, predicting adherence to medications \citep{yin_therapy_2018}, predicting hospital admissions \citep{handly_evaluation_2015, zhang_prediction_2017}, autosuggest for documentation improvement \citep{jernite_predicting_2013}, and patient portal message classification \citep{cronin_automated_2015, cronin_comparison_2017, sulieman_classifying_2017}. Obtaining chief complaints is also critical for the development of novel use cases, such as automatic patient smart triage and virtual care for low complexity cases. 

Historically, utilizing chief complaint data has been challenging due to low rates of patient health literacy, high rates of misspelled words \citep{shapiro_taming_2004}, and variance in how patients describe symptoms \citep{jones_patients_2010, may_emergency_2010, raven_comparison_2013}. Previous approaches to chief complaint identification relied heavily on n-gram models trained with classical machine learning techniques \citep{brown_ngram_2010, chapman_classifying_2005, chapman_classification_2005, jernite_predicting_2013, cronin_automated_2015, cronin_comparison_2017}. More recently, convolutional neural networks (CNNs) \citep{sulieman_classifying_2017} and recurrent neural networks \citep{lee_chief_2018} have been employed. An important limiting factor to the use of neural networks in extracting chief complaints are the relatively small sizes of available annotated datasets, which makes learning from scratch difficult. 

In this work, we present a new dataset of patient free-text entries that have been matched to structured chief complaints chosen by intake nurses at the beginning of corresponding patient encounters. We compare several free-text embedding models based on fine-tuned Bidirectional Transformers (BERT) \citep{devlin_bert:_2019}, as well as provide a baseline TF-IDF embedding trained on the patient text. We propose a simple model that can combine these embeddings with discrete data, such as patient age and sex, and explore model performance with additional misspellings added to the test set. Finally, we perform a qualitative evaluation using experimenter-derived queries to show the impact of context, demographic data, and common misspellings. 

\section{Chief Complaint Dataset}

We developed a new dataset, Chief Complaint (CC) Dataset, consisting of patient-generated text, discrete chief complaints, and demographic information. The dataset was collected from Kaiser Permanente (KP) patients in Southern California who booked a Primary Care Physician appointment through our web portal, kp.org, between February 2018 and June 2019. Patients were prompted to “Tell us more about what you need from this appointment" and then provided with a text-field constrained to 50 characters for their response. Subsequently, at the time of check-in for their appointment, an intake nurse would choose appropriate discrete chief complaints with the goal of categorizing the patient's reason-for-visit as accurately as possible. For our analysis, the reason-for-visit, discrete chief complaint, patient age and biological sex information were all extracted from KP's Electronic Medical  Record (EMR). 

Of the 323,903 encounters retrieved, there were 277,297 unique text entries that were assigned 474,012 chief complaints (of which 1,851 were unique). A subset of encounters had very generic patient text or non-specific chief complaints, and we used simple string matching to remove such examples (\nameref{section:appendixA}). The resulting dataset consisted of 222,058 encounters for 168,328 members. There were 180,562 unique free-text entries mapped to a total of 282,636 discrete chief complaints, 1,701 of which were unique. For our model development, we chose to classify only those chief complaints that appear at least 10 times, which left us with 795 unique chief complaints.  The median reason for visit text was 38 characters long, with a median word count of 7 (5 without stop words).  For the reasons-for-visit entries with additional free-text comments added by intake personnel, we cropped out comments to enforce a 50 character limit.

Among the patients in the final chief complaint dataset, 92,848 (55\%) were female while 75,476 (45\%) were male. The average age was 49 years old. A full breakdown by age and sex can be found in~\nameref{section:appendixB}, Tables \ref{table:datasetSexDistribution} and \ref{table:datasetAgeDistribution}.

The dataset was split into a training set of 190,243 encounters and 31,815 test encounters such that no sentences in the test set had an exact match in the training set.  Out of the test set, we selected 245 unique patient reasons-for-visit entries that contained at least one clear misspelling. The test and misspelling sets were used only in the final evaluation of the models.

\section{Model}
\label{section:model}

A feed-forward neural network was constructed to combine pre-trained free-text embeddings with patient age (discretized into age groups) and biological sex (see \nameref{section:appendixC}, Fig.~\ref{figure:ccmodel4}). As multiple discrete chief complaints can be extracted from a given patient-generated reason for visit, the output of the model is a one-vs-rest classification for 797 chief complaints. 

Pre-trained models for text embeddings were obtained from three publicly available sources: BERT-base \citep{devlin_bert:_2019}, BioBERT \citep{lee_biobert:_2019}, and Clinical BERT \citep{alsentzer_publicly_2019}. In addition, we fine-tuned a KP BERT model starting with Clinical BERT, using both the masked language model and the next sentence prediction task with 100 million sentence pairs extracted from patient progress notes (from encounters unrelated to the Chief Complaint Dataset). We also fine-tuned Chief Complaint (CC) BERT on the training portion of the CC Dataset, starting from KP BERT, using the masked language model prediction task. To generate BERT-based embeddings, we mean-pooled the last 4 layers, yielding an embedding size of 3,072. 

A baseline embedding model using TF-IDF was also trained on the training portion of the CC Dataset. Pre-processing before TF-IDF vectorization included: (1) transforming the text to lowercase; (2) removing standard English language stop-words; (3) replacing all numbers with the generic “\#” symbol. The Penn Treebank (PTB3) tokenizer  implemented in NLTK \citep{bird_natural_2009} was used to tokenize the text, and unigrams and bigrams were then computed. TF-IDF vectors were created consisting of values for the 50,000 most common n-grams, with at most 0.5 document frequency in the training set. The vectors were normalized to unity using the l2-norm.

To incorporate demographic data, embeddings were learned directly and sized to be the same as the text embeddings (3,072 for BERT based embeddings and 50,000 for TF-IDF embeddings). The demographic embeddings were then additively combined with the pre-trained free-text embeddings. 

The final embeddings were fed into a  dense ReLU layer before being passed to the output layer. The output layer consisted of one-vs-rest logistic regressions for 795 discrete chief complaints. Dropout was used after the embedding layers and after the ReLU layer to regularize the network. Models were trained using binary cross-entropy loss and the Adam optimizer \citep{kingma_adam:_2017} with an initial learning rate of 0.001 and batch size of 1024.  Validation precision-recall area under the curve (PR-AUC) was used to guide the learning rate schedule with a patience of 10 epochs and a learning rate factor of 0.8. Early stopping was used if validation PR-AUC did not improve for 20 epochs, which typically happened after 40-50 epochs of training.

We tested the following model hyperparameters: (1) hidden layer sizes $\in {500, 1000, 2000}$ (2) dropout rates  $\in {0, 0.2, 0.3}$. For each hyperparameter combination, we performed  5-fold cross validation to generate multiple validation, test, and misspelling metric estimates. For each embedding type we selected the layer-size and dropout-rate which produced the best average validation PR-AUC. For all of the BERT embedding based models the hidden layer size was 500 and the dropout rate was 0.3. For the TF-IDF embedding based model the hidden layer size was 2000 and the dropout rate was also 0.3. 

\section{Results}
\label{section:results}

We built models that predicted one-vs-rest probability for 795 chief complaints that had at least 50 occurrences in the training set. Language embeddings were based on CC BERT, KP BERT, Clinical BERT, BioBERT, BERT-base, and TF-IDF. For each model, we generated 5 estimates by training in 5-fold cross validation. Resulting average and standard deviation micro PR-AUC and micro ROC-AUC values can be found in Table~\ref{table:testSetPerformance}.

\begin{table}[h!]
	\caption{Test and misspelling set performance using various text embeddings}
	\centering
	\begin{tabular}{lllll}
	\toprule
                  & \multicolumn{2}{c}{\textbf{Test set}}  & \multicolumn{2}{c}{\textbf{Misspellings set}} \\
    \midrule
                & \textbf{PR AUC}        & \textbf{ROC AUC}       & \textbf{PR AUC}           & \textbf{ROC AUC}           \\
    \toprule
BERT base       & 0.3439±0.0038 & 0.9692±0.0008 & 0.2482±0.0101    & 0.9544±0.0038    \\
BioBERT         & 0.3400±0.0044 & 0.9689±0.0008 & 0.2361±0.0043    & 0.9529±0.0028     \\
Clinical BERT   & 0.3458±0.0048 & 0.9690±0.0012 & 0.2384±0.0048    & 0.9554±0.0028     \\
KP BERT         & 0.3567±0.0078 & \textbf{0.9700±0.0015} & 0.2470±0.0101    & 0.9557±0.0022     \\
CC BERT & \textbf{0.3597±0.0041} & 0.9691±0.0011 & \textbf{0.2579±0.0079}    & \textbf{0.9576±0.0005}     \\
\midrule
TF-IDF baseline & 0.3878±0.0148 & 0.9762±0.0016 & 0.2733±0.0130    & 0.9548±0.0030    \\
\bottomrule

		\label{table:testSetPerformance}
	\end{tabular}
\end{table}

To compare statistical significance, we performed a Welch's t-test using micro PR AUC. All of the BERT models performed similarly (best BERT performance highlighted in bold in Table~\ref{table:testSetPerformance}), although there is a statistically significant improvement between Clinical BERT and KP BERT ($p=0.006$ for test set, and $p=0.025$ for misspelling set). An unexpected outcome, however, is that the TF-IDF embeddings performed better than even the CC Dataset fine-tuned BERT (CC BERT), with $p=7\cdot 10^{-5}$ on the test set, and a statistically not significant $p=0.06$ on the misspelling set. 

\begin{figure}[!h]
	\centering
	\includegraphics[width=1\linewidth]{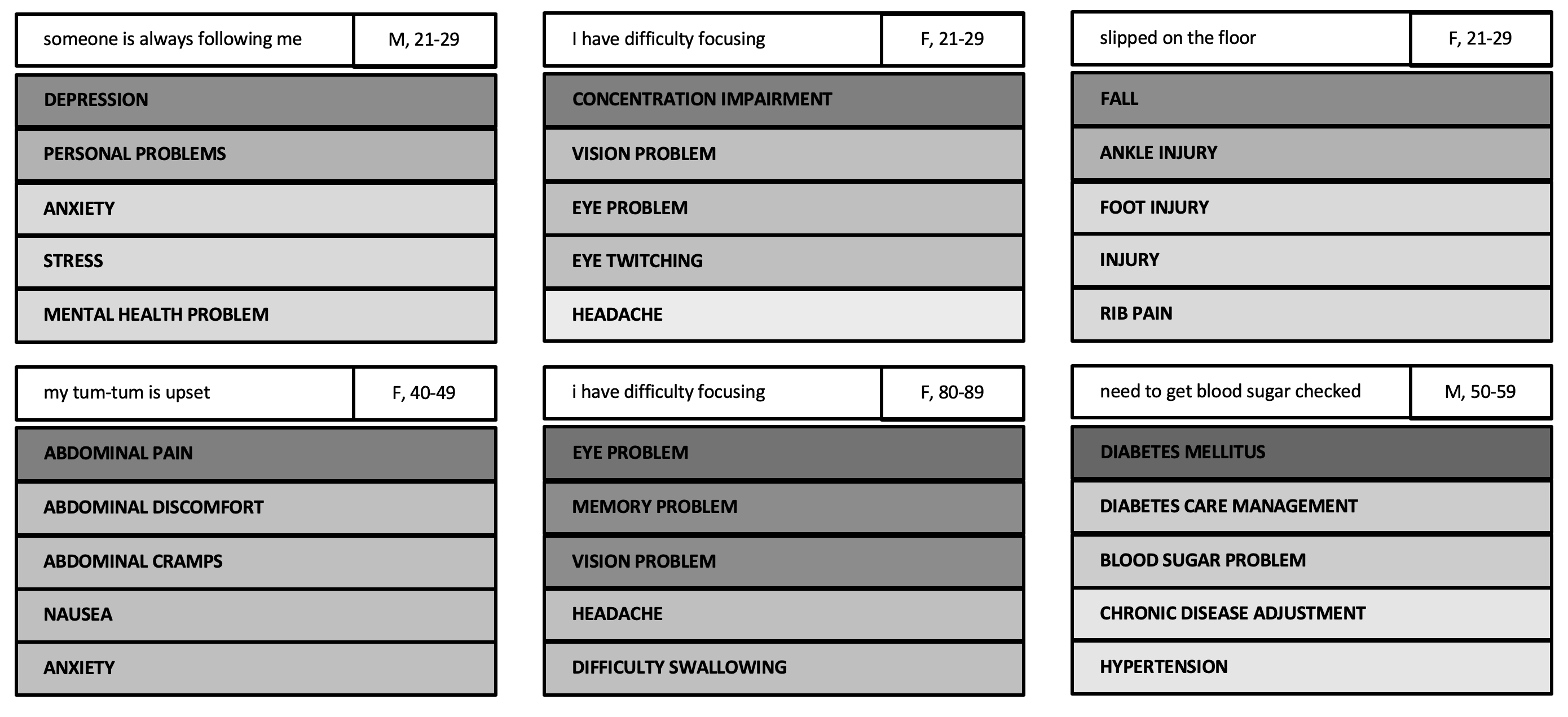}
	\caption{Example queries not from the dataset as predicted with the KP BERT text embeddings. Color intensity corresponds to the relative prediction weight.}
	\label{figure:exampleQueries}
\end{figure}

Examples of queries and responses (experimenter-generated, not found in the Chief Complaint Dataset) with our KP BERT model can be seen in Fig.~\ref{figure:exampleQueries}. The darkness of the gray color represents relative weight of the prediction. There is successful identification of chief complaints from queries with rare or colloquial words such as “my tum-tum is upset,” which returns abdominal pain as the number one structured complaint. The model also shows sensitivity to demographic data, with the query “I have difficulty focusing” returning different top structured complaints for younger patients (“concentration impairment” for ages 21-29) compared to older patients (“eye problem” and “memory problem” for ages 80-89).

We further compared classification TF-IDF based model with KP BERT based model using experimenter-generated queries in (\nameref{section:appendixE}). Our results show that while TF-IDF model performs strongly on the patient-generated data, it may be not as robust as BERT based models to unusual or highly misspelled entries.

\section{Conclusions}
\label{section:conclusions}

We compared the performance of five BERT models with baseline TF-IDF embeddings on the important clinical natural language understanding task of extracting chief complaints from patient generated text. To do this we developed a new dataset consisting of patient complaints in their own words, mapped structured complaints chosen by intake personnel, and patient demographics. We built a small feed-forward model that combined the pre-trained embeddings with learned embeddings for demographics to enable utilization of these heterogeneous data sources, further improving accuracy when predicting structured complaints. 
 
Given the relatively small size of the Chief Complaint Dataset, we were initially surprised that the baseline TF-IDF embedding model performed better ($p<7\cdot 10^{-5}$) than fine-tuned BERT embedding models in micro PR AUC statistics. The good performance of TF-IDF is likely due in part to the fact that after stop word removal the median free text entry was only 5 tokens long. Thus, bags of n-grams were able to capture some of the context and sequencing of words. While our current dataset is limited to small and simple reasons-for-visit, we did observe that for more complicated experimenter-generated queries BERT-based models seemed more robust.
 
Future work will involve expanding the Chief Complaint Dataset and re-mapping the structured complaints to a more disjoint set. In this work, we only collected a small amount of patient specific, structured data (age and sex). It would be fruitful to include additional patient data (e.g. medication history and diagnosis history) extracted from the EMR to provide a broader context for patient complaints. We also plan to increase the allowed length of the patient reason-for-visit entry to enable more complex free-text entries. Finally, we plan to explore transformer-based models that will be fine-tuned on more patient text data sources, such as patient emails to clinical staff. We expect to better understand the robustness of these models as we seek direct patient and clinician feedback as part of a planned virtualized-clinical-triage pilot.

\bibliographystyle{ieeetr}
\bibliography{cc2discrete.bbl}

\clearpage

\appendix
\section*{Appendix A}
\label{section:appendixA}

\renewcommand{\thesubsection}{A.\arabic{subsection}}
\renewcommand{\thetable}{A\arabic{table}}

\setcounter{table}{0}
\setcounter{subsection}{0}

\subsection{Dataset token exclusions}
\label{subsection:datasetTokenExcl}

Many encounters had very generic patient-entered free-text, and were coded to non-disease state specific discrete chief complaints. We removed encounters whose discrete chief complaints contained only those including the following tokens: REFILL, IMMUNIZATION, PHYSICAL, FOLLOW UP,  FLU SHOT, TELEPHONE, ERROR, CHRONIC CONDITION, MEDICATION, LAB, MAINTENANCE, RESULTS, APPOINTMENT, EXAM, COUNSELING, REVIEW, TELECARE, GOALS OF CARE, LOW ADHERENCE. A total of 150 unique discrete chief complaints contained these tokens.

\clearpage

\section*{Appendix B}
\label{section:appendixB}

\renewcommand{\thesubsection}{B.\arabic{subsection}}
\renewcommand{\thetable}{B\arabic{table}}

\setcounter{table}{0}
\setcounter{subsection}{0}

\subsection{Dataset distribution details}
\label{subsection:datasetDistribution}

\begin{table}[!h]
	
	\caption{Sex distribution of patients in the dataset. Percentage from total in parenthesis}
	\centering
	\begin{tabular}{lll}
		\toprule
		\textbf{Biological Sex} & \textbf{Number of patients} & \textbf{Number of encounters} \\
		\toprule
		Female                  & 92,848 (55.2\%)             & 112,758 (55.4\%)              \\
		Male                    & 75,476 (44.8\%)             & 90,845 (44.6\%)               \\
		\bottomrule
		\label{table:datasetSexDistribution}
	\end{tabular}
\end{table}

\begin{table}[!h]
	\caption{Age distribution of patients in the dataset. Percentage from total in parenthesis.}
	\centering
	\begin{tabular}{lll}
		\toprule
		\textbf{Age Group (years old)} & \textbf{Number of patients} & \textbf{Number of encounters} \\
		\toprule
		0-1                            & 885 (0.5\%)                 & 1,122 (0.5\%)                 \\
		2-15                           & 8,514 (5.0\%)               & 10,253 (5.0\%)                \\
		16-20                          & 1,817 (1.1\%)               & 2,060 (1.0\%)                 \\
		21-29                          & 20,907 (12.4\%)             & 24,326 (11.9\%)               \\
		30-39                          & 28,148 (16.7\%)             & 32,833 (16.1\%)               \\
		40-49                          & 23,795 (14.1\%)             & 28,440 (14.0\%)               \\
		50-59                          & 27,939 (16.6\%)             & 33,770 (16.6\%)               \\
		60-69                          & 31,429 (18.7\%)             & 39,037 (19.2\%)               \\
		70-79                          & 18,219 (10.8\%)             & 23,253 (11.0\%)               \\
		80-89                          & 5,528 (3.3\%)               & 7,047 (3.5\%)                 \\
		90+                            & 1,147 (0.7\%)               & 1,467 (0.7\%)                 \\
		\bottomrule
		\label{table:datasetAgeDistribution}
	\end{tabular}
\end{table}

\clearpage

\section*{Appendix C}
\label{section:appendixC}

\renewcommand{\thesubsection}{C.\arabic{subsection}}
\renewcommand{\thefigure}{C\arabic{figure}}

\setcounter{figure}{0}
\setcounter{subsection}{0}

\begin{figure}[!h]
	\centering
	\includegraphics[width=0.6\linewidth]{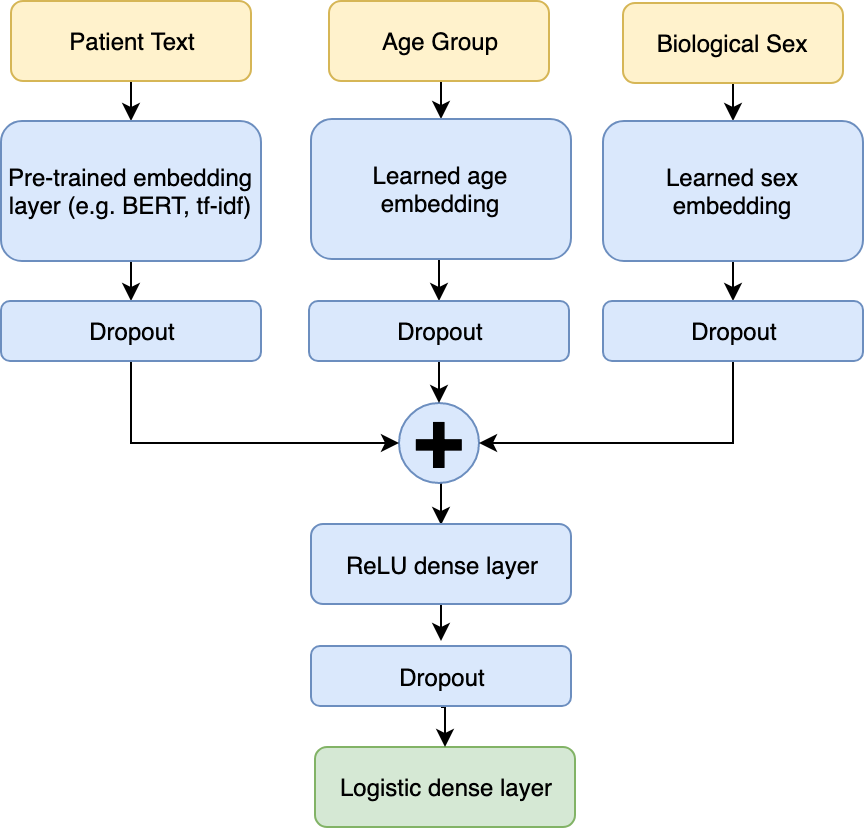}
	\caption{Schematic of the free-text to Discrete Chief Complaint model. Inputs are marked in yellow, internal layers are light blue, and the output layer is light green. Pre-trained embeddings were used for patient text while age-group and biological sex demographic embeddings were trained. All of the embeddings were then summed and passed to a ReLU activated dense layer. The output layer was a one-vs-rest logistic dense layer. }
	\label{figure:ccmodel4}
\end{figure}

\clearpage

\section*{Appendix D}
\label{section:appendixD}

\renewcommand{\thesubsection}{D.\arabic{subsection}}
\renewcommand{\thefigure}{D\arabic{figure}}

\setcounter{figure}{0}
\setcounter{subsection}{0}

\subsection{Clustering chief complaints}
\label{subsection:ccClustering}

\begin{figure}[hb!]
	\centering
	\begin{subfigure}[b]{0.49\linewidth}
		\includegraphics[width=\linewidth]{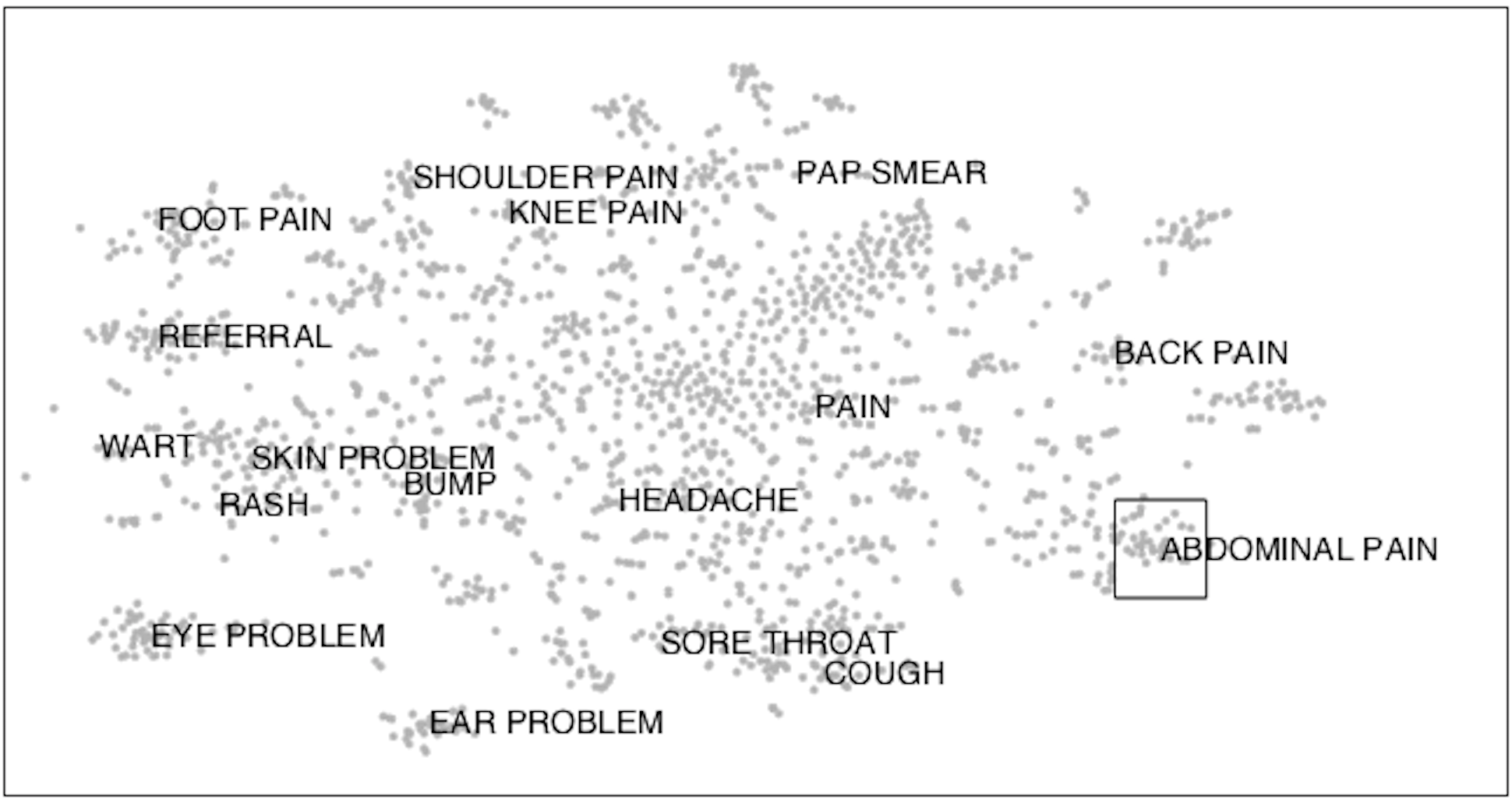}
		\caption{All of the chief complaints}
	\end{subfigure}
	\begin{subfigure}[b]{0.49\linewidth}
		\includegraphics[width=\linewidth]{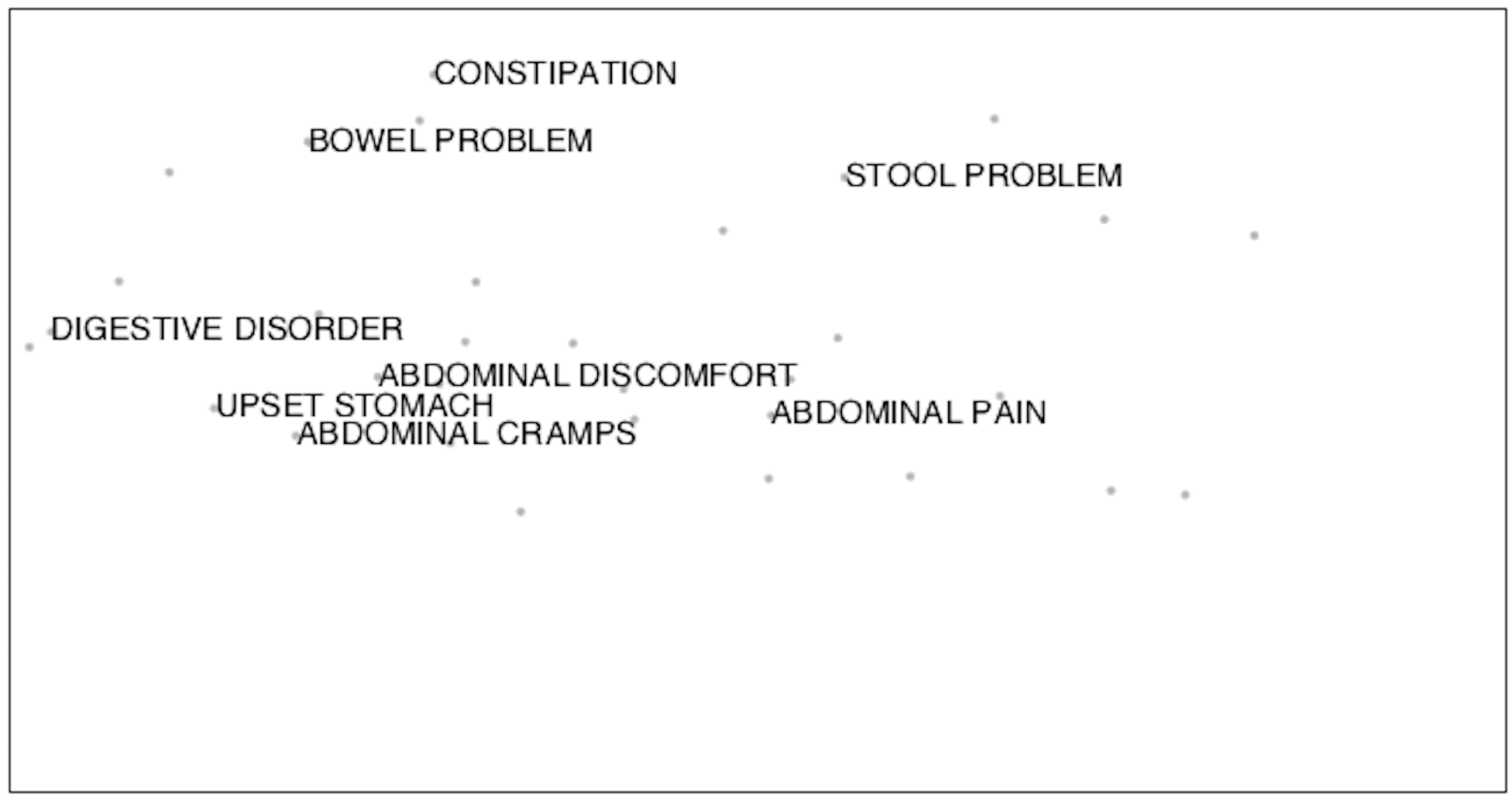}
		\caption{Blow up of the rectangle from (a)}
	\end{subfigure}
	\caption{ t-SNE clustering of average discrete chief complaint probability vectors. Related complaints appear close to each other.}
	\label{figure:tSNE}
\end{figure}

We performed a clustering of the discrete chief complaints based on reason-for-visit classification model. For each chief complaint, we computed the mean predicted chief complaint vector. A t-SNE \citep{maaten_visualizing_2008} 2D projection of the mean predicted chief complaint vectors is shown in Fig.~\ref{figure:tSNE}a. Each dot represents a discrete chief complaint, which naturally cluster by anatomical region and problem type. High frequency representatives of each cluster are annotated. Fig.~\ref{figure:tSNE}b focuses on the region directly surrounding the “abdominal pain” chief complaint. All 25 discrete chief complaints in this area are related to the abdominal region.

\clearpage

\section*{Appendix E}
\label{section:appendixE}

\renewcommand{\thesubsection}{E.\arabic{subsection}}
\renewcommand{\thetable}{E\arabic{table}}

\setcounter{table}{0}
\setcounter{subsection}{0}

\subsection{Evaluation using experimenter-generated queries}
\label{subsection:evaluationExperimenterGen}

Experimenter-generated queries were used to further test model sensitivity to misspellings  (examples in Table~\ref{table:modelSensitivityExperimenterQueries}). We compared KP BERT and TF-IDF models without demographic data, as these models demonstrated similar performance on the test set (Table 1). Frequently, KP BERT's top suggestions were robust to human-typical misspellings. However, the TF-IDF model tended to predict unrelated but high marginal probability chief complaints (COUGH, BACK PAIN, etc.) when key words were altered.

\begin{table}[!h]
	\centering
	\caption{Examples of top chief complaint (CC) predictions from the two language models without demographics on experimenter created free-text entries with spelling errors.}
	\begin{tabular}{lll}
		\toprule
		\textbf{ Text}                    & \textbf{ KP BERT Top CC} & \textbf{TF-IDF Top CC} \\
		\toprule
		i have had a headche for two days & HEADACHE                 & COUGH                  \\
		my visaon is bad                  & EYE PROBLEM              & COUGH                  \\
		stomic pain                       & ABDOMINAL PAIN           & BACK PAIN              \\
		i am dizzzy a lot                 & DIZZY                    & COUGH                  \\
		feel like i can't brethe          & SHORTNESS OF BREATH      & COUGH                  \\
		\bottomrule
		\label{table:modelSensitivityExperimenterQueries}
	\end{tabular}
\end{table}

Many of the patient-generated complaints in the Chief Complaint Dataset were formulaic and rather simple in nature, leading us to explore model performance with experimenter-generated, context-heavy free-text complaints. Table~\ref{table:modelClassificationsExperimenterQueries} shows example top classifications for the KP BERT and the TF-IDF based, no demographic models. Often, KP BERT was able to correctly predict queries that were context-heavy, while the TF-IDF based model either selected a high marginal probability chief complaint or latched onto a strong indicating token (“discomfort whenever I move” mapped to “abdominal discomfort”).

\begin{table}[!h]
	\caption{Examples of top chief complaint (CC) predictions from the two language models without demographics on experimenter created free-text entries.}
	\begin{tabular}{lll}
		\toprule
		\textbf{ Text}                   & \textbf{KP BERT Top CC}      & \textbf{TF-IDF Top CC}        \\
		\toprule
		my ibd is acting up                 & ABDOMINAL PAIN      & KNEE PROBLEM         \\
		i take a lot of naps                & SLEEP PROBLEM       & COUGH                \\
		i don't work well with others & DEPRESSION & TELEMEDICINE \\ & & CONSULT \\
		the world appears hazy              & VISION PROBLEM      & COUGH                \\
		my face hurts all the time          & HEADACHE            & BACK PAIN            \\
		i feel wobbly                       & DIZZY               & COUGH                \\
		i feel upset every day              & ANXIETY             & HEADACHE             \\
		discomfort whenever I move    & PAIN       & ABDOMINAL    \\ & & DISCOMFORT \\
		the concert last night was too loud & EAR PROBLEM         & COUGH                \\
		i worked out too hard at the gym    & BACK PAIN           & COUGH                \\
		i lie awake in bed at night         & SLEEP PROBLEM       & FATIGUE              \\
		slipped on the ice                  & FALL                & BACK PAIN            \\
		feel like i can't get enough oxygen & SHORTNESS OF BREATH & COUGH                \\
		i feel tingly all over              & TINGLING            & DIZZY               \\
		\bottomrule
		\label{table:modelClassificationsExperimenterQueries}
	\end{tabular}
\end{table}

\end{document}